\documentclass[lettersize,journal]{IEEEtran}
\newcommand{\etal}{\textit{et al.}}
\usepackage{amsmath,amsfonts}
\usepackage{algorithmic}
\usepackage{algorithm}
\usepackage{array}
\usepackage[caption=false,font=normalsize]{subfig}
\usepackage{textcomp}
\usepackage{stfloats}
\usepackage{url}
\usepackage{verbatim}
\usepackage{graphicx}
\usepackage{cite}
\usepackage{multirow}
\usepackage{makecell}
\usepackage{tabularx}
\usepackage{xspace}
\usepackage{booktabs}
\usepackage{cellspace}
\usepackage{capt-of}
\usepackage{colortbl}
\usepackage{pdfpages}
\usepackage[commandnameprefix=ifneeded]{changes}
    \definechangesauthor[name=Martin,color=green]{m}
\usepackage{xcolor}
\begin{document}

\title{Code and Pixels: Multi-Modal Contrastive Pre-training for Enhanced Tabular Data Analysis}

\author{Kankana Roy, \IEEEmembership{Member, IEEE}, Lars Krämer, 
	Sebastian Domaschke,
    Malik Haris, 
	Roland Aydin, 
	Fabian Isensee, and
	Martin Held
\thanks{Kankana Roy is with the Karolinska Institute, Solnavagen 1, Stockholm, Sweden.
  	    (e-mail:kankana.kankana.roy@gmail.com)\\
  	    Lars Krämer and Fabian Isensee are with Helmholtz Imaging and the German Cancer Research Center (DKFZ),  Heidelberg, Germany
  	    (e-mail: f.isensee@dkfz-heidelberg.de; lars.kraemer@dkfz-heidelberg.de)\\
  	    Sebastian Domaschke and Roland Aydin are with Institute of Materials Systems Modeling, Helmholtz-Zentrum Hereon, Geesthacht, Germany 
  	    (e-mail: Domaschke@mail.com; roland.aydin@hereon.de)\\
        Malik Haris and Martin Held are with Institute of Membrane Research,  Helmholtz-Zentrum Hereon, Geesthacht, Germany
  	    (e-mail: malik.haris@hereon.de; martin.held@hereon.de)
  	    }
        }

\markboth{Journal of \LaTeX\ Class Files,~Vol.~14, No.~8, August~2021}%
{Shell \MakeLowercase{\textit{et al.}}: A Sample Article Using IEEEtran.cls for IEEE Journals}
%

\maketitle

\begin{abstract}
Learning from tabular data is of paramount importance, as it complements the conventional analysis of image and video data by providing a rich source of structured information that is often critical for comprehensive understanding and decision-making processes. We present Multi-task Contrastive Masked Tabular Modeling (MT-CMTM), a novel method aiming to enhance tabular models by leveraging the correlation between tabular data and corresponding images. MT-CMTM employs a dual strategy combining contrastive learning with masked tabular modeling, optimizing the synergy between these data modalities.

Central to our approach is a 1D Convolutional Neural Network with residual connections and an attention mechanism (1D-ResNet-CBAM), designed to efficiently process tabular data without relying on images. This enables MT-CMTM to handle purely tabular data for downstream tasks, eliminating the need for potentially costly image acquisition and processing.

We evaluated MT-CMTM on the DVM car dataset, which is uniquely suited for this particular scenario, and the newly developed HIPMP dataset, which connects membrane fabrication parameters with image data. Our MT-CMTM model outperforms the proposed tabular 1D-ResNet-CBAM, which is trained from scratch, achieving a relative $1.48\%$ improvement in relative MSE on HIPMP and a $2.38\%$ increase in absolute accuracy on DVM. These results demonstrate MT-CMTM's robustness and its potential to advance the field of multi-modal learning.

Code and data will be made publicly available upon acceptance.
\end{abstract}

\begin{IEEEkeywords}
masked tabular modeling, multi-modal contrastive learning, 1D-ResNet-CBAM, multi-modal fusion
\end{IEEEkeywords}

\begin{figure*}[th]
	\centering
	\includegraphics[width=\textwidth]{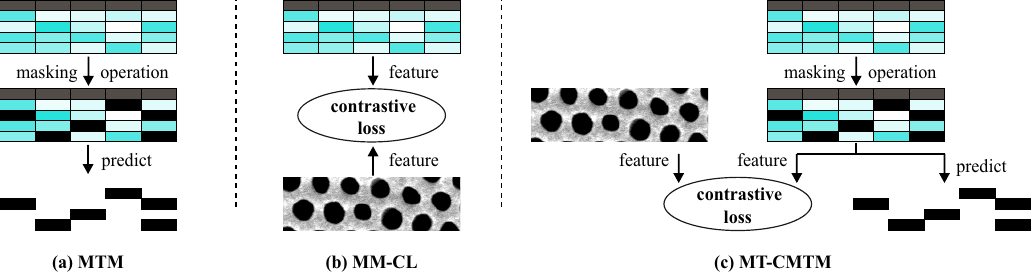}
	\caption{Connecting two pretext tasks used in self-supervised tabular pre-training: (a)~Masked Tabular Data Modeling (MTM) conducts a mask-and-predict pretext task. (b)~Multi-modal contrastive learning (MM-CL) follows a modality comparison paradigm. (c)~Multi-task Contrastive Masked Tabular Modeling (MT-CMTM) introduces a combined contrastive and masked pretext scheme.}
	\label{fig:feature-graphic}
\end{figure*}
\section{Introduction}
\label{sec:intro}

In traditional computer vision, the focus on images stems from the cost-effectiveness of their acquisition, especially for natural images with abundant, freely available databases~\cite{deng2009imagenet}. The surge in machine learning has broadened the scope of image-processing algorithms, making them more versatile. This integration has driven computer vision into mainstream research, necessitating adaptation across diverse domains.

However, this paradigm shift presents unique challenges, especially in areas where the cost of capturing images is significant. In our problem context, tabular data is abundant, while generating imaging data is resource-intensive. Hence, we would like to build a tabular model that is enriched with auxiliary image modality. The case of relying solely on tabular data for independent predictions without corresponding images has not been the focus of our research, which results in very limited availability of validation data for our goal to establish a robust image-enriched tabular model. While several datasets integrating both image and tabular modalities have been documented, most do not match our case or are not publicly accessible. Some datasets focus on image data of documents but lack the text version~\cite{Appalaraju.DocFormerv2,Huang.LayoutLMv3}. Medical datasets come with specific intricacies, such as having to use 3- instead of 2-dimensional image data to accurately represent the details, as is the case for the ADNI dataset, linking MRI scans to tables of medical history and genetic markers~\cite{Huang.ADNI,Petersen.ADNI}. Often, medical datasets lack a strong tabular modality to be used as an independent input without imaging modality, as in the OLIVES datsaset~\cite{prabhushankar2022olives}. Datasets of consumer products may be very sparse with incomplete tabular data not a very good input for tabular model, e.g. the M5product dataset~\cite{Dong.M5Product}. The DVM dataset~\cite{hager2023best,huang2022dvm} stands out as an exception, being publicly available, dense and aligning with our research needs. The high-dimensional tabular data in the DVM dataset enables independent predictions without images, making it suitable for our objectives and the only existing option for bench marking.

In parallel, we introduce the new Hereon Isoporous Polymer Membrane Production dataset (HIPMP dataset), combining membrane manufacturing process parameters and their corresponding images. The dataset treats isoporous block-copolymer membranes, offering a uniform pore structure for cost-effective water filtration to tackle global clean water challenges~\cite{Abetz.2015}. Despite 14 years of research, understanding how the fabrication parameters affect pore morphology remains incomplete~\cite{Muller.2021}. While some parameters, like polymer concentration, can be controlled, others, such as starting material properties, pose challenges~\cite{Radjabian.2020}. Due to interdependencies and kinetic factors, the discovery of optimal fabrication parameters requires trial and error to date~\cite{Rangou.2014}. This highlights the need for enhanced control and reproducibility in membrane production. A predictive model, leveraging fabrication parameters as tabular input, membrane micrographs as auxiliary image data, and morphological membrane quality parameters as tabular output would be an essential step in solving this pressing issue.

A common approach is to use a joint multi-modal image+tabular model to leverage both modalities. However, our specific goal is to develop a predictive tabular model using only tabular and auxiliary imaging modalities, under the consideration that images are only accessible during the training phase. It's essential to note that existing tabular data modeling techniques may lack the adaptability to seamlessly integrate the imaging modality during training~\cite{chen2016xgboost,popov2019neural,arik2021tabnet}.

We introduce a novel pre-training strategy that enriches the tabular model for a downstream task where only tabular data is accessible. 
This strategy employs a self-supervision approach in which masked tabular modeling (MTM) involves training on pretext tasks (Fig. \ref{fig:feature-graphic}a), such as predicting masked regions from corrupted inputs~\cite{yoon2020vime,he2022masked}. The MTM model comprises two main components: an encoder that handles corrupted data and a decoder responsible for predicting the correct data. Our primary focus lies on the encoder because it is used for downstream tasks. However, the MTM strategy can sometimes negatively impact the encoder's performance due to the coupled decoder. Recent research, exemplified in~\cite{mao2022improvements}, suggests that simplifying the decoder can be beneficial. We posit that by minimizing the decoder's influence while retaining its predictive capabilities, thereby enhancing the encoder's performance.
The MTM loss function imposes a common constraint on both the encoder and decoder, lacking an independent loss function solely for the encoder. Consequently, to enhance the encoder's feature extraction capabilities, we introduce new tasks into the framework.

Another concurrent concept is multi-modal contrastive learning (MM-CL) (Fig. \ref{fig:feature-graphic}b), where the tabular model is compelled to align with corresponding image features in a shared feature space~\cite{chen2020simple}. 

By employing MM-CL as a pre-training with MTM, the proposed Multi-task Contrastive Masked Tabular Modeling (MT-CMTM) (Fig. \ref{fig:feature-graphic}c) adopts a multi-task approach. Introducing two tasks is expected to augment the encoder's performance in downstream applications of MT-CMTM.

In brief, our work brings key contributions to the field:
\begin{itemize}
	\item \textbf{HIPMP Dataset:} We introduce HIPMP, a meticulously curated dataset that provides a comprehensive overview of isoporous block-copolymer membrane production. This dataset includes all relevant fabrication parameters, accompanying images, and high-level morphological quality descriptors. It serves as an invaluable resource for the development of multi-modal and inverse machine learning techniques, opening up exciting possibilities for research and innovation.
	\item \textbf{1D-ResNet-CBAM Tabular Encoder:} We present a novel network architecture, 1D-ResNet-CBAM, designed specifically for tabular data. This encoder sets a strong benchmark for modeling tabular data, showcasing its potential for enhancing data analysis and prediction tasks.
	\item \textbf{Enhancing Performance with MT-CMTM:} To demonstrate the effectiveness of our approach, we conduct experiments on two distinct tabular data modeling problems, utilizing the HIPMP and DVM~\cite{huang2022dvm} datasets. Our experimental analysis indicates that the MT-CMTM approach can improve the 1D-ResNet-CBAM model's performance in our test scenarios, highlighting its potential in real-world applications.
\end{itemize}

\section{Related Work}
\subsection{Self-supervised Learning in Tabular Data}
Self-supervised frameworks aim to acquire data representations through unlabeled data and can be broadly categorized into two types: those utilizing pretext tasks and those employing contrastive learning. Most existing self-supervised frameworks with pretext tasks are designed primarily for image and natural language modalities. For example, they include tasks such as image recovery from randomly masked input~\cite{he2022masked,mao2022improvements}, predicting RGB values of raw pixels~\cite{xie2022simmim}, masked representation prediction~\cite{chen2023context}, and calculating perceptual similarity~\cite{dong2023peco}.

In the context of tabular data, previous studies have explored self-supervised learning as well. For instance, the denoising auto-encoder~\cite{vincent2008extracting} focuses on restoring the original sample from a corrupted version. Similarly, the context encoder~\cite{chen2023context} employs a pretext task involving the reconstruction of the original sample using both the corrupted sample and a mask vector. Self-supervised learning in TabNet~\cite{arik2021tabnet} and TaBERT~\cite{yin2020tabert} also revolves around the recovery of corrupted tabular data as their designated pretext task. Another notable contribution comes from VIME~\cite{yoon2020vime}, which introduces the novel pretext task of mask vector recovery, serving as inspiration for our tabular self-supervised prediction task.

\subsection{Multi-modal Contrastive Learning}
The introduction of SimCLR~\cite{chen2020simple} has popularized contrastive learning, extending into various multi-modal tasks~\cite{guo2023pace,liu2023multimodal}, including video-audio contrastive learning~\cite{zhao2022self,haliassos2022leveraging,assefa2023audio}, video-text contrastive learning~\cite{luo2021coco}, image-natural language contrastive learning~\cite{yuan2021multimodal,mustafa2022multimodal}, and medical images-genetic data~\cite{taleb2022contig}.

Despite this trend, limited exploration has occurred in modeling images and tabular data contrastively. Recent works have begun to show interest in this area. In~\cite{dong2022m5product}, a unified multi-modal representation is learned from various modalities, including tabular data, considering incomplete and noisy data. Both~\cite{hager2023best,huang2023multimodal} focus on multi-modal contrastive learning in medical imaging, where patient data typically involves image and tabular pairs.

However, our problem differs from these works. In membrane research, images are generated post-production for quality assessment, making this feedback mechanism costly. Thus, it would be advantageous if the model could predict quality exclusively from tabular data without relying on images for the end task.

\subsection{Multi-task Learning for Self-supervision}
Multi-task learning, with a well-established history in computer vision, aims to improve encoder generalizability and task performance~\cite{misra2016cross,kendall2018multi,kanaci2019multi,liu2019self}. In contrastive learning, some approaches use it as an auxiliary task based on images to enhance primary task performance in a multi-task framework~\cite{li2021multi,neto2021focusface,yin2022contrastive,zhang2023multi}.

Our method, in contrast, employs multi-modal contrastive learning and a self-supervised pretext task in pre-training without integrating them into the main task. A similar work we are aware of is~\cite{mao2022improvements}, which is entirely image-based, combining two subtasks for pre-training.

\section{Preliminaries}
\label{sec:method}

\begin{figure*}
	\includegraphics[width=\textwidth]{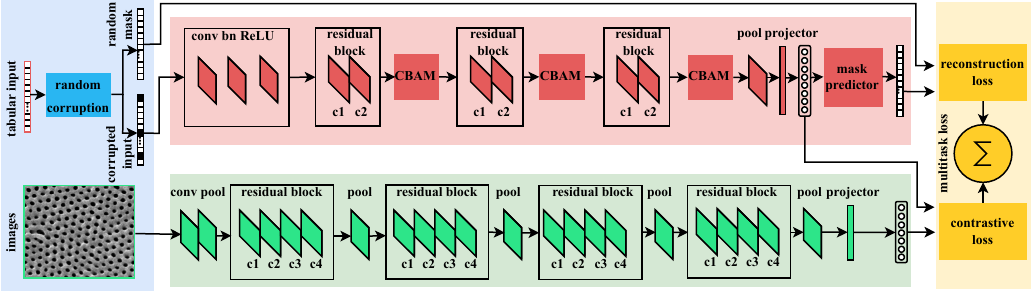}
	\caption{Incorporating self-supervised pretext tasks and multi-modal contrastive learning in a multi-task framework. Notably, the unimodal tabular encoder (\textit{highlighted in red}) exhibits substantial performance improvements when shared between two distinct pre-training strategies.}
	\label{fig:2}
\end{figure*}

\subsection{Masked Tabular Modeling (MTM)}
\label{sec:mtm}
The core objective of MTM is to optimize a pretext model for mask vector estimation, given a corrupted input sample. Initially, a mask vector generator produces a binary mask vector $m = [m_1,...,m_l]^T \in {0,1}^l$, with $m_j$ sampled from a Bernoulli distribution characterized by probability $p_m$. Subsequently, a pretext generator $g_m : \mathcal{X} \times {0,1}^l \rightarrow \mathcal{X}$, operating on a sample $x_t$ from dataset $D$ and mask vector $m$, generates a masked sample $\tilde{x}_t$ following the equation:
\begin{equation}
	\tilde{x}_t = g_m(x_t,m) = m \odot \bar{x}_t + (1 - m) \odot x_t
	\label{eq:4}
\end{equation}
where the the $j$-th feature of $\bar{x}_t$ is derived from the empirical distribution, denoted as $\hat{p}(x_j)$. This empirical distribution is calculated by averaging over the presence of each unique value $x_{i,j}$ in the $j$-th feature of all samples within the dataset $D$. Essentially, it represents the empirical marginal distribution for each individual feature. This process ensures that the corrupted sample $\tilde{x}_t$ retains tabular characteristics while resembling the samples in $D$. The randomness introduced by the Bernoulli-distributed vector $m$ and the stochastic nature of the pretext generator $g_m$ (derived from $\bar{x}_t$) collectively amplify the difficulty in reconstructing $x_t$ from $\tilde{x}_t$. The difficulty level is adjustable via the hyperparameter $p_m$, controlling the proportion of features masked and corrupted.

The mask vector estimator, denoted as $s_m: \mathcal{Z} \rightarrow [0,1]^l$, takes $z_t$ and predicts a vector $\hat{m}$ that identifies features in $\tilde{x}_t$ replaced by noisy counterparts (i.e., $m$).

The primary loss function $L_m$ is the sum of $L_1$ losses for each dimension of the mask vector:
\begin{equation}
	L_1(m, \hat{m}) = \frac{1}{d} \sum_{j=1}^l |m_j - (s_m \circ E_t)_j(\tilde{x}_t)|
	\label{eq:5}
\end{equation}
Here, $\hat{m} = (s_m \circ E_t)(\tilde{x}_t)$.

\subsection{Multi-modal Contrastive Learning (MM-CL)}
\label{sec:mmcl}
Multi-modal contrastive learning transfers knowledge from an unlabeled dataset to a labeled task, by leveraging multiple modalities associated with the same data point. Our focus is to pre-train a tabular model using paired images and tabular data in dataset $D$. Each $x \in D$ comprises an image $x^i \in \mathbb{R}^{H\times W \times C}$ and tabular data $x^t \in \mathbb{R}^{L}$.

For our multi-modal contrastive learning model, separate encoders for images and tabular data are employed. The tabular data encoding employs 1D-ResNet-CBAM (supplementary material) with convolutional layers. The image encoder utilizes a ResNet-18~\cite{he2016deep} backbone with global average pooling and an MLP projection head.

In the contrastive learning framework, InfoNCE loss~\cite{oord2018representation} distinguishes positive image-tabular feature pairs. Image encoder $E_i$ transforms an image into $v_i$, and tabular encoder $E_t$ transforms masked tabular data $\tilde{x}_t$ into $\tilde{v}_t$. These are projected into a common space via MLP heads, producing $z_i$ and $\tilde{z}_t$. The loss, as in Eq. \eqref{eq:1}, Eq. \eqref{eq:2}, involves treating features as queries and keys, generating $L_{it}$ and $L_{ti}$. The final contrastive loss, Eq. \eqref{eq:3}, distinguishes and leverages paired image-tabular features in line with contrastive learning principles.

\begin{equation}
	L_{it} = -\log \frac{\exp (z_i.\tilde{z}_t^+/\tau)}{\exp (z_i.\tilde{z}_t^+/\tau) + \sum{\tilde{z}_t}-\exp (z_i.\tilde{z}_t^-/\tau)}
	\label{eq:1}
\end{equation}

\begin{equation}
	L_{ti} = -\log \frac{\exp (\tilde{z}_t.z_i^+/\tau)}{\exp (\tilde{z}_t.z_i^+/\tau) + \sum{z_i}-\exp (\tilde{z}_t.z_i^-/\tau)}
	\label{eq:2}
\end{equation}

\begin{equation}
	L_c = \frac{1}{2}L_{it} + \frac{1}{2}L_{ti}
	\label{eq:3}
\end{equation}

\section{Multi-task Contrastive Masked Tabular Modeling (MT-CMTM)}
Our MT-CMTM builds upon the advancements of MTM, aiming to harness the full potential of the encoder in the context of self-supervised learning. Illustrated in Fig. \ref{fig:2}, our comprehensive framework comprises three pivotal components:
\begin{enumerate}
	\item \textbf{Mask Vector Prediction Pretext Task:}
	To endow the MTM encoder with attribute information about tabular data, we devise the mask vector prediction task. This task involves corrupting tabular data and providing the MT-CMTM encoder with masked tabular data for effective feature extraction.
	\item \textbf{Contrastive Tasks:} We introduce a multi-modal contrastive task tailored for the MT-CMTM tabular encoder with masked data. This novelty enhances the encoder's feature extraction capabilities through contrastive learning, allowing the MT-CMTM encoder to capture distinctive features intrinsic to tabular data. This augmentation, in turn, bolsters the overall performance of MTM during downstream tasks.
	\item \textbf{Multi-task Training MT-CMTM Encoder:}
	Embracing a multi-task training paradigm, we concurrently optimize the MT-CMTM encoder for both the multi-modal contrastive task and the mask vector estimation pretext task. This simultaneous training approach, with a shared tabular encoder, facilitates the reinforcement of the tabular encoder by leveraging knowledge from the imaging modality. This strategic fusion enhances the overall robustness and knowledge transfer capabilities of the MT-CMTM framework.
\end{enumerate}

In this formulation, consider a multi-task scenario with two tasks, each associated with an individual loss function $L_c$ and $L_m$ and learnable weight parameters $\lambda_c$ and $\lambda_m$. The overall loss function $L$ is expressed as a linear combination of the individual losses, where the weights are adaptively adjusted during training:
\begin{equation}
	L = \lambda_c L_c + \lambda_m L_m
	\label{eq:6}
\end{equation}

\section{Experiments and Results}
\label{exp}

\subsection{Hereon Isoporous Polymer Membrane Production (HIPMP) Dataset}
\begin{table}[t]
	\centering
	\caption{Datasets attributes}
	\scalebox{1.1}{
		\begin{tabular}{c|c|c}
			\hline
			Attribute & \makecell{DVM\\dataset} & \makecell{HIPMP\\dataset}\\\hline
			Number of images & 176,414 & 1970\\\hline
			Train:validation:test & 70,565:17,642:88,207 & 1262:315:393\\\hline
			Tabular input size & 14 & 19 \\\hline
			Categorical feature & 4 & 3\\\hline
			Morphometric feat. & 5 & 0 \\\hline
			Number of output & 286 & 33 \\\hline
	\end{tabular}}
	\label{tab:dataattri}
\end{table}

The HIPMP dataset is a comprehensive collection of electron microscopy images of the surface pore structures of PS-P4VP copolymer membranes, the extracted membrane morphological quality properties from them, and their corresponding fabrication parameters. This dataset enables relating the fabrication parameters with the morphology properties of the resulting membrane to tailor insights into the membrane fabrication process.

The membrane morphology properties were extracted via a semantic segmentation map of characteristic membrane features which is then used to extract the descriptors. Initially, classes of interest like different types of pores or membrane defects were defined by human experts. In an iterative manner, images were labeled, and deep-learning semantic segmentation models~\cite{Isensee.2021}  were trained. By inspecting and validating the model predictions the most informative unlabeled images were selected, labeled, and added to the training pool for the next iteration. After the results were verified as sufficient by the human experts, segmentation masks were generated for each image using the final model. These masks were used to extract membrane morphology properties such as pore size, pore circularity, or pore distribution.

The result is a comprehensive database that contains 1970 grayscale images (input) of size $1024 \times 708$~pixels where each is described by 19 fabrication parameters as input and 33 membrane morphology properties (33 non-sparse out of 47) as output which is condensed in Tab. \ref{tab:dataattri}. As this dataset contains unpublished values, a smaller public version was constructed, as well (more detail in the supplementary materials)~\cite{HIPMP}. 

\subsection{Data Visual Marketing (DVM) Dataset}
\label{dataset}
The DVM dataset, sourced from $335,562$ pre-owned car advertisements~\cite{huang2022dvm}, contains $1,451,784$ car images taken from various angles (45-degree increments), along with sales and technical information. To evaluate our approach's validity and generalization capabilities, we conducted a task similar to Hager~\etal~\cite{hager2023best}: predicting car models from images and associated advertising data. 
We followed Hager~\etal's experimental setup~\cite{hager2023best}, including 14 input details like width, length, height, wheelbase, price, year, mileage, seating capacity, door count, original price, engine capacity, body style, transmission type, and fuel category. Notably, brand and model year information was excluded, and we made random adjustments to the size measurements by up to 50 millimeters to prevent unique identification. We combined this tabular data with a random image input from each advertisement, creating a dataset with $70,565$ training pairs, $17,642$ validation pairs, and $88,207$ test pairs. We excluded car models with fewer than 100 samples, resulting in a total of 286 target classes (output).

To handle missing data in our datasets, we applied the mean value interpolation method, a useful technique for data imputation. This method involves estimating and filling missing values using information from nearby data points. By applying this method, we ensured that our datasets remained suitable for analysis, with minimal disruption to the overall data structure and integrity. 

\subsection{Implementation Details}
Our framework was implemented using the PyTorch deep learning library, and we utilized the Nvidia GeForce RTX 4090 graphics card for both training and testing. During the pre-training phase, we conducted training on the HIPMP dataset for 25 epochs, while the DVM dataset underwent training for 10 epochs. Subsequently, in the downstream task, we further trained both the HIPMP and DVM datasets for 100 epochs and 10 and 552 steps for HIPMP and DVM datasets respectively.
For both training phases, we employed the Adam Optimizer with a learning rate of $1\times10^{-2}$ and a weight decay of $0.0001$. Additionally, we incorporated the OneCycleLR optimizer, following the methodology proposed by Smith~\etal~\cite{smith2019super}.
For data preprocessing, we applied a center crop of size $224\times224$~pixels for the DVM dataset, while a random crop was used for the HIPMP dataset. Our dataset split strategy involved a 5-fold cross-validation for the HIPMP dataset, and the split ratios specified in Tab. \ref{tab:dataattri}. For the DVM dataset, we followed the training, validation, and testing split recommended in the work of Hager~\etal~\cite{hager2023best}.

\subsection{Metrics}
In this study, we aim to evaluate performance enhancements in comparison to state-of-the-art tabular methods and our self-defined benchmark (Tab. \ref{tab:realtedwork}). For the regression task using the HIPMP dataset, we input fabrication parameters and predict a 33-dimensional morphological quality metric vector, assessed through mean absolute error (MAE) and mean squared error (MSE). For the classification task with the DVM dataset, we use tabular features associated with advertisements to predict their classification labels. Key classification metric accuracy, balanced accuracy, and mean F-score are the primary evaluation criterion here. 

We also analyze the model's computational characteristics, including the number of model parameters, floating-point operations (FLOPs), and training overhead, as detailed in Tab. \ref{tab:model_burden}.

Two ablation analyses by varying the number of layers in the proposed 1D-ResNet-CBAM to find the optimal architecture suitable for our problem and another study by varying the number of samples in training and how it affects the performance of the proposed MT-CMTM can be found in the Sec. \ref{sec:ablation_varying_layers} and Sec. \ref{sec:ablation_low_data}. As expected, increasing the number of layers and samples increases the performance. Notably, MT-CMTM outperforms other models for the regression task even at very small datasets.

\subsection{Results and Comparison with the Related Work}
\begin{table*}[ht]
	\centering
	\caption{Performance of our MT-CMTM framework on the tasks of membrane quality metric prediction and DVM car model prediction from images. The best-performing model is displayed in bold font.}
	\scalebox{1.2}{
		\begin{tabular}{c|l|ccc|cc}
			\hline
			&\multirow{2}{*}{Model}& \multicolumn{3}{|c}{\textbf{DVM dataset}} 
			& \multicolumn{2}{|c}{\textbf{HIPMP dataset}}\\\cline{3-7}
			&  & Mean Acc. $\uparrow$&\shortstack{Bal.\\Acc.} $\uparrow$& \shortstack{Mean\\F-score} $\uparrow$& MAE$\downarrow$ & MSE$\downarrow$ \\\hline
			\multirow{4}{*}{\shortstack{Classical\\models}} 
			& Linear model~\cite{freedman2009statistical} & 4.40  & 0.89 & 0.134& 0.356$\pm.001$ &  0.511$\pm.005$ \\
			& Ridge model~\cite{hilt1977ridge} & 10.99 & 3.01 & 0.041 & 0.357$\pm.002$ &   0.512$\pm.003$ \\
			& SVM model~\cite{cortes1995support} & 1.74 & 0.90 & 0.008 & 0.298$\pm.004$ &  0.554$\pm.007$ \\
			& XGBoost~\cite{chen2016xgboost} & 72.60 & 63.90 & 0.709 & 0.203$\pm.001$ &   0.523$\pm.004$ \\ \hline
			\multirow{2}{*}{\shortstack{Transformer\\models}} 
			&TabNet~\cite{arik2021tabnet} & 87.51 & 85.28 & 0.872 & 0.242$\pm.005$ &  0.844$\pm.008$ \\
			& 1D-Transformer~\cite{vaswani2017attention} & 54.81  & 43.69 & 0.498 & 0.231$\pm.003$ &  0.476$\pm.007$ \\
			\hline
			1D deep learn. models
			&1D-CNN~\cite{baosenguo2021} & 60.09  & 58.67 & 0.580 & 0.208$\pm.004$ &  0.433$\pm.011$ \\
			
			\hline
			\multirow{4}{*}{Ours}& 1D-ResNet-CBAM (PM) & 89.73  & 88.24 & 0.895 & 0.180$\pm.000$ &  0.372$\pm.004$ \\
			& PM + MT-CMTM & \textbf{92.11}  & 90.86 & 0.920 & \textbf{0.178}$\pm.003$ & \textbf{0.366}$\pm.010$ \\
            & \makecell[l]{PM + MT-CMTM\\on public HIPMP} & n.a. & n.a. & n.a. & 0.187$\pm.003$ & 0.651$\pm.006$\\
			\hline
	\end{tabular}}
	\label{tab:realtedwork}
\end{table*}

We conduct a comparative analysis of our approach with established tabular models, including the Linear model~\cite{freedman2009statistical}, Ridge model~\cite{hilt1977ridge}, support vector machine (SVM)~\cite{cortes1995support}, and the decision forest-based XGBoost~\cite{chen2016xgboost}. In the context of tabular data modeling, we also benchmark our method against transformer-based models, specifically the 1D-transformer inspired by Vaswani~\etal~\cite{vaswani2017attention} and TabNet~\cite{arik2021tabnet}. Moving on to our proposed model, we employ a 1D-CNN architecture~\cite{baosenguo2021} and further enhance it with 1D-ResNet-CBAM, drawing inspiration from residual connections~\cite{he2016deep} and and convolutional block attention modules~\cite{woo2018cbam}. The detailed architecture of this proposed model can be found in the supplementary. In a related recent study~\cite{hager2023best}, an attempt was made to leverage tabular data for modeling image modality, presenting a counterpart to our problem setting. To provide insights, we present experimental results on the DVM dataset, focusing on tabular data only. 
Tab. \ref{tab:realtedwork} presents the experimental results obtained on both the DVM and HIPMP datasets.

Classical models, including Linear regression~\cite{freedman2009statistical}, Ridge regression~\cite{hilt1977ridge}, and SVM~\cite{cortes1995support}, exhibit notably poor performance. These models face significant challenges in handling the complex correlations within the dataset, and the intricacies of the classification task, leading to their inability to effectively capture the input-output relationships.
In contrast, the recently introduced XGBoost~\cite{chen2016xgboost}, a leading model for tabular data, demonstrates satisfactory performance with an accuracy of $72.6\%$, a balanced accuracy of $63.9\%$ and a mean F-score of $0.71$ for the DVM dataset, while for  HIPMP it delivered an MSE of $0.52$ and MAE of $0.20$.

When modeling a 1D-transformer architecture inspired by Vaswani~\etal~\cite{vaswani2017attention}, our results did not yield a substantial improvement.
Conversely, TabNet, a transformer-based tabular model proposed by Arik~\etal~\cite{arik2021tabnet}, exhibited strong performance in the case of DVM, achieving an accuracy of $87.5\%$, balanced accuracy of $85.3\%$ and a mean F-score of $0.87$, but a poorer MSE of $0.84$ and MAE of $0.24$ for the HIPMP dataset.

We employed a specific variant of 1D-CNN, inspired by the work of Baosenguo~\etal~\cite{baosenguo2021}, commonly used in tabular data modeling. However, similar to the standard 1D-CNN, this variant did not yield remarkable results when applied to the DVM dataset.
In contrast, our proposed 1D-ResNet-CBAM architecture produced remarkable outcomes, achieving an accuracy of $89.7\%$, a balanced accuracy of $88.2\%$ and a mean score of $0.90$ for the DVM data, as well as an MSE of $0.37$ and MAE of $0.18$ for the HIPMP case. This achievement serves as our main benchmark, highlighting the effectiveness of this model in the context of our study.

We conducted a comparison with Hager's image-based method~\cite{hager2023best} on the DVM dataset. Hager \etal~\cite{hager2023best} reported a top-1 accuracy superior to MT-CMTM by a margin of $2.29\%$. However, it's important to consider that their method relies on image data, which is a more powerful modality compared to our tabular-only approach in this specific application. Therefore, the observed performance gain can be attributed, in part, to the use of a stronger modality.

In our final results, after pre-training and fine-tuning for the downstream task, our model achieved an impressive accuracy of $92.1\%$, balanced accuracy of $90.9\%$ and a mean F-score of $0.92$ and an MSE of $0.37$ and an MAE of $0.18$. That constitutes an absolute enhancement of $2.4\%$ in accuracy and a relative improvement of $1.5\%$ in MSE compared to the proposed 1D-ResNet-CBAM benchmark. Evidently, the multi-modal contrastive pre-training in MT-CMTM benefits tabular data analysis in both classification and regression tasks.

For the 24\% smaller, public version of the HIPMP dataset~\cite{HIPMP}, our MT-CMTM model achieves a slightly higher MAE of $0.19$ than the full-size dataset, while the MSE of $0.65$ is significantly higher. This finding foreshadows the insights of Sec.~\ref{sec:ablation_low_data} and trend in Fig. \ref{fig:sampleNr}, where a reduction of the dataset size increases the error of the model. This outcome is even enhanced by the fact that the public dataset contains a higher fraction of membranes with a non-uniform pore morphology, creating (relatively) more outliers, thus increasing the MSE disproportionately compared to MAE. Nonetheless, these additional results underline the model's robust performance.

\subsection{Ablation analysis of different training and pre-training strategies}
\label{sec:trainingstrat}
\begin{table*}[h!]
	\caption{Ablation performance analysis with different training and pre-training strategies on DVM and HIPMP datasets.}
	\centering
	\scalebox{1.3}{
		\begin{tabular}{l|ccc|cc}
			\hline
			\multirow{2}{*}{Experiment} &\multicolumn{3}{|c|}{\textbf{DVM dataset}} &\multicolumn{2}{|c}{\textbf{HIPMP dataset}}\\\cline{2-6}
			& \shortstack{Mean\\Accuracy}$\uparrow$ &\shortstack{Balanced\\Accuracy}$\uparrow$ &\shortstack{Mean\\F-score}$\uparrow$ &MAE $\downarrow$ &MSE $\downarrow$\\\hline 
			1D-ResNet-CBAM (PM)  & 89.73 & 88.24 & 0.895 & 0.180$\pm.000$ & 0.372$\pm.004$\\
			PM + pretext mask    & 90.79 & 89.22 & 0.900 & 0.178$\pm.001$ & 0.370$\pm.004$\\
			PM + pretext feature & 89.73 & 88.24 & 0.895 & 0.180$\pm.003$ & 0.372$\pm.008$\\
			PM + MM-CL           & 89.82 & 88.49 & 0.897 & 0.180$\pm.002$ & 0.371$\pm.003$\\
			PM + MT-CMTM         & 92.11 & 90.23 & 0.916 & 0.178$\pm.003$ & 0.366$\pm.010$
			\\\hline
	\end{tabular}}
	\label{tab:ablationstudies}
\end{table*}

In order to delve into the underlying reasons behind the enhanced performance of unimodal encoders through pre-training via a pretext task in a multi-task fashion, we conducted a comprehensive analysis of the distinct contributions made by various pre-training strategies. These findings are presented in Tab. \ref{tab:ablationstudies}. We focus on the impact on the DVM dataset, as the metrics of the HIPMP dataset vary merely within the margin of error.

Our investigation commences with the proposed model (PM), 1D-ResNet-CBAM, which serves as our initial reference point. In this configuration, no supplementary training strategies are employed, and only the tabular modality is utilized to train the model from scratch. The proposed model's performance is detailed in the first row of Tab. \ref{tab:ablationstudies}, achieving an accuracy of $89.73\%$, a balanced accuracy of $88.24\%$ and a mean F-score of $0.895$ for the DVM dataset.

In a bid to draw inspiration from the masked image modeling pre-training paradigm~\cite{he2022masked}, we embarked on a similar approach tailored for tabular models. In this strategy, the input tabular data is deliberately corrupted using a random mask, and the pretext task involves either estimating the mask vector or reconstructing the feature vector. Subsequently, the encoder and output estimator are trained to utilize a self-supervised loss function (Fig. \ref{fig:feature-graphic}a). The knowledge gained from this training phase is then employed to initialize the 1D-ResNet-CBAM model for downstream tasks.
The results stemming from the first pretext task, which revolves around mask vector estimation, are presented in the second row of Tab. \ref{tab:ablationstudies}. Encouragingly, we observe a small but notable improvement in both the classification and regression tasks, signifying the efficacy of this pre-training strategy. The mask vector estimation strategy resulted in an accuracy of $90.79\%$, a balanced accuracy of $89.22\%$ and a mean F-score of $0.900$ for the DVM dataset. 

In the third row of the table, we present the outcomes of the feature vector reconstruction task from corrupted input. Regrettably, this particular pretext task did not yield a discernible impact on the subsequent downstream tasks. For the DVM dataset, the feature vector reconstruction strategy achieved an accuracy of $89.73\%$, a balanced accuracy of $88.24\%$ and a mean F-score of $0.895$.

Furthermore, we explored the untapped potential of the image modality, even when not directly used for downstream tasks. Inspired by~\cite{hager2023best}, we adopted a multi-modal contrastive learning approach to maximize mutual information between modalities relating to the same data point (Fig. \ref{fig:feature-graphic}b).
We used the pre-trained ResNet-18 architecture~\cite{he2016deep} on the ImageNet-1K dataset~\cite{deng2009imagenet} as the image encoder. We initialized our downstream model with the weights learned by the 1D-ResNet-CBAM encoder.
The results, presented in Tab. \ref{tab:ablationstudies} as PM + MM-CL, demonstrate a performance boost compared to the proposed model, showcasing the effectiveness of our multi-modal contrastive learning strategy in harnessing image modality information for improved results. This approach resulted in an accuracy of $89.82\%$, a balanced accuracy of $88.49\%$ and a mean F-score of $0.897$ for the DVM dataset.

To boost model performance, we merged our two most promising approaches into a weighted multi-task model.
These two approaches include the self-supervised pretext task of mask vector prediction as in Sec. \ref{sec:mtm} (Fig. \ref{fig:feature-graphic}a), and the image-tabular modality-based MM-CL, as outlined in Sec. \ref{sec:mmcl} (Fig. \ref{fig:feature-graphic}b).
The culmination of these efforts, presented in Tab. \ref{tab:ablationstudies} as PM + MT-CMTM (Fig. \ref{fig:feature-graphic}c), has delivered the best results so far: a competitive accuracy of $92.11\%$, balanced accuracy of $90.23\%$ and mean F-score of $0.916$ for the DVM dataset. These improved metrics establish the weighted multi-task model as the top-performing strategy.

\subsection{Complexity Analysis}
\begin{table*}[t]
	\centering
	\caption{Trade-off between the number of parameters, FLOPs, and MSE were calculated over the HIPMP dataset.}
	\scalebox{1.3}{
		\begin{tabular}{c|c|c|c|c}
			\hline
			\multirow{2}{*}{Network} & Parameters & FLOPs & Training & \multirow{2}{*}{MSE} 
			\\&(million) &(million) & strategy &\\\hline
			TabNet~\cite{arik2021tabnet} & 0.55 & 2.19 &  main task & 0.888\\
			1D Transf.~\cite{vaswani2017attention} & 0.052 & 0.27 & main task & 0.476\\
			1D-CNN~\cite{baosenguo2021} & 3.56 & 32.1 & main task & 0.433\\
			1D-ResNet-CBAM (PM) & 1.79 & 7.47 & main task & 0.372\\
			PM + MT-CMTM & 12.98 & 51.6 & pre-train. & 0.366\\\hline
	\end{tabular}}
	\label{tab:model_burden}
\end{table*}

Tab. \ref{tab:model_burden} provides a detailed comparison of these models, emphasizing key metrics such as the number of parameters, FLOPs (floating-point operations), and MSE. The evaluation is conducted on the HIPMP dataset, encompassing different baselines and training strategies.
In the MT-CMTM, the entries only consider the pre-training cost, with the downstream task cost staying the same as in the previous 1D-ResNet-CBAM row. This suggests a potential efficiency gain in predictive performance, even with higher training computational cost.

\subsection{Ablation Analysis of Losses}
\begin{table*}[t]
	\centering
	\caption{Illuminating the effect of loss functions on model performance: The first seven rows delve into pre-training, while the last six rows scrutinize downstream tasks. Loss abbreviations: CE (Cross-Entropy), BCE (Binary Cross-Entropy), BT (Barlow Twin), HL (Huber Loss), FL (Focal Loss).}
	\scalebox{1.3}{
		\begin{tabular}{c|c|c|c}
			\hline
			\multirow{2}{*}{\makecell{Module\\and phase}} & \multirow{2}{*}{Loss} &  \textbf{DVM}& \textbf{HIPMP}\\\cline{3-4}
			&  & Mean Acc. $\uparrow$&   MSE $\downarrow$\\\hline
			\multirow{3}{*}{\makecell{MTM \textcolor{gray}{+ 1D-ResNet-CBAM}\\(pre-training)}} & MSE & 88.61 & 0.368$\pm.007$\\
			& $L_1$ & 90.52 & 0.368$\pm.005$\\
			& CE & 87.82 & 0.367$\pm.004$\\\hline
			\multirow{4}{*}{\makecell{MM-CL \textcolor{gray}{+ 1D-ResNet-CBAM}\\(pre-training)}} & InfoNCE~\cite{oord2018representation} & 91.59 & 0.367$\pm.004$\\
			&CLIP~\cite{radford2021learning} & 86.30 & 0.369$\pm.003$\\
			&SimSiam~\cite{chen2021exploring} & 91.55 & 0.368$\pm.005$\\
			&BT~\cite{zbontar2021barlow} & 90.87 & 0.376$\pm.007$\\\hline\hline
			\multirow{3}{*}{\makecell{\textcolor{gray}{MT-CMTM +} 1D-ResNet-CBAM\\(downstream regression)}}& MSE & n.a. & 0.480$\pm0.56$\\
			& $L_1$ & n.a. & 0.364$\pm.004$\\
			& HL & n.a. & 0.364$\pm.004$\\\hline
			\multirow{3}{*}{\makecell{\textcolor{gray}{MT-CMTM +} 1D-ResNet-CBAM\\(downstream classification)}}& CE & 91.59 & n.a.\\
			& BCE & 89.40 & n.a.\\
			& FL & 85.22 & n.a.\\\hline
	\end{tabular}}
	\label{tab:lossablation}
\end{table*}

To determine the optimal loss function for our training strategies, we conducted an ablation study for pre-training and downstream tasks (Tab. \ref{tab:lossablation}). For the MTM (Fig. \ref{fig:feature-graphic}a) pre-training strategy, keeping the downstream fixed, we experimented with MSE, $L_1$, and cross-entropy (CE) loss. We found that $L_1$ and CE loss outperformed MSE, and we selected $L_1$ loss for MT-CMTM training as later we noticed it performed optimally compared to CE.

Next, for MM-CL (Fig. \ref{fig:feature-graphic}b), we tested various contrastive loss functions in the pre-training with fixed downstream, including InfoNCE~\cite{oord2018representation}, CLIP~\cite{radford2021learning}, SimSiam~\cite{chen2021exploring}, and Barlow Twin (BT~\cite{zbontar2021barlow}). We observed that InfoNCE performed competitively, even when compared to the popular multi-modal contrastive loss CLIP for this specific task.

After pre-training, setting the best loss for both and combining them into MT-CMTM, we transitioned to the downstream task, where we needed suitable loss functions. For regression, we tested MSE, $L_1$, and Huber loss, with $L_1$ and Huber performing well. For classification, we explored CE, balanced CE (BCE), and focal loss (FL)~\cite{lin2017focal}, with CE proving to be superior. 

In summary, $L_1$ loss was used for the multi-task segment in MT-CMTM, InfoNCE loss for the contrastive learning segment, while $L_1$ loss was applied for regression downstream tasks, and CE loss for classification.

\subsection{Ablation Study Varying 1D-ResNet-CBAM Layers}
\label{sec:ablation_varying_layers}
\begin{figure}[h!]
	\centering
	\subfloat[DVM]{\includegraphics[width=0.25\textwidth]{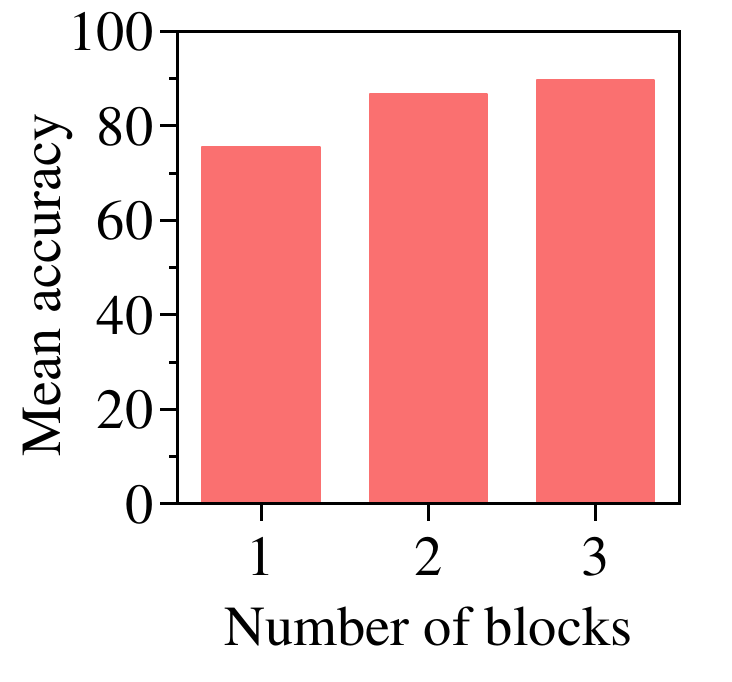}\label{fig:layerNr_DVM}}
	\subfloat[HIPMP]{\includegraphics[width=0.25\textwidth]{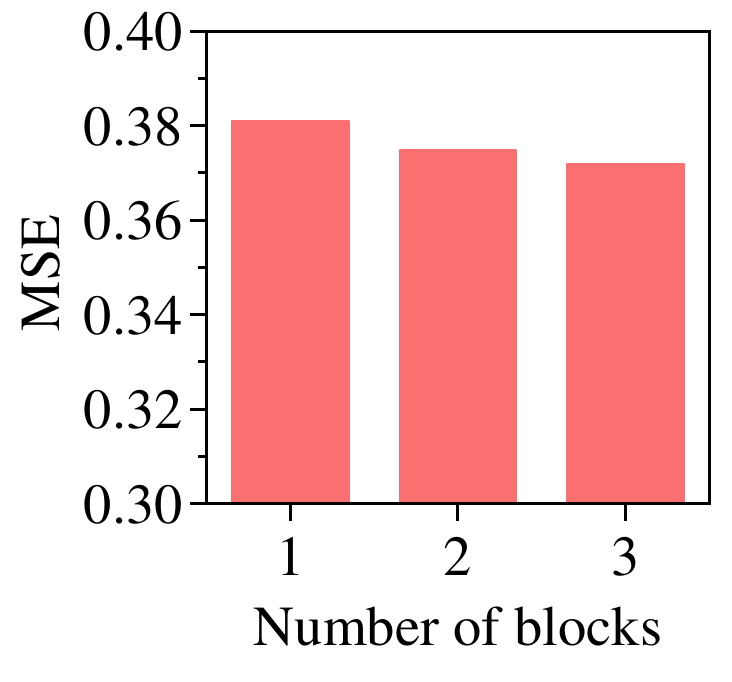}\label{fig:layerNr_XIPMP}}
	\caption{Effect of the number of convolution blocks on the performance of our proposed model 1D-ResNet-CBAM  on the tasks of (a) DVM car model prediction from images and (b) membrane quality metric prediction.}
	\label{fig:layerNr}
\end{figure}

In the ablation study conducted to investigate the impact of varying the number of layers (convolution blocks in red followed by a CBAM block in Fig. 1 of the supplementary materials) in the 1D-ResNet-CBAM architecture, we systematically examined models with different block configurations as shown in Fig. \ref{fig:layerNr}. Our findings revealed that the choice of the number of blocks significantly influences the performance of the network. Specifically, increasing the number of blocks led to improved model performance, higher mean accuracy or lower MSE, with a clear correlation between the network's depth and its ability to capture complex features and patterns. 

\begin{figure}[h!]
	\centering
 	\subfloat[DVM]{\includegraphics[width=0.24\textwidth]{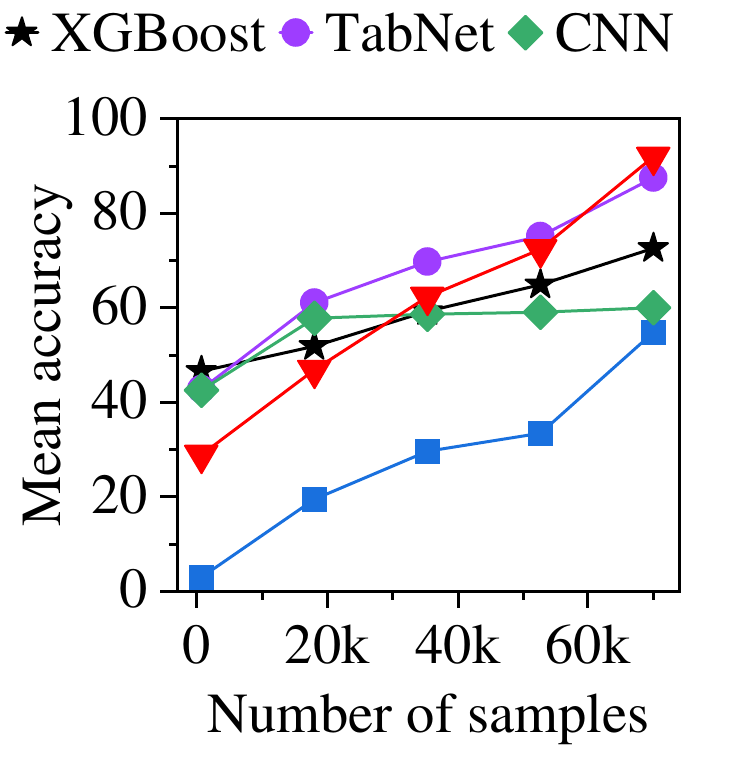}\label{fig:sampleNr_DVM}}\hfill
	\subfloat[HIPMP]{\includegraphics[width=0.24\textwidth]{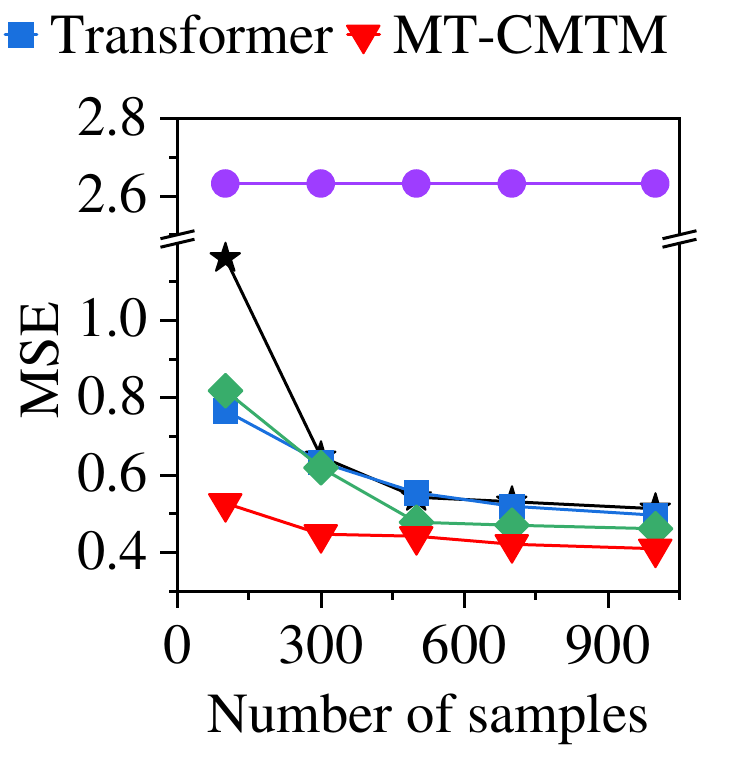}\label{fig:sampleNr_HIPMP}}
	\caption{Effect of the number of samples on the performance of our proposed MT-CMTM model versus tabular models on the tasks of (a) DVM car model prediction from images and (b) membrane quality metric prediction.}
	\label{fig:sampleNr}
\end{figure}

\subsection{Ablation Study in Low Data Regimes}
\label{sec:ablation_low_data}
To assess the performance of our learned encoders in a low-data scenario, we systematically sampled the fine-tuning training dataset at intervals from 100\% down to 10\%. The results are graphically represented in Fig. \ref{fig:sampleNr}. Our model consistently outperforms other tabular models, especially in the case of the HIPMP dataset, even in extremely low-data scenarios. While for DVM, our model is initially outperformed by other tabular models in the first two quarters (700 to 14,000 data points), its performance ranks highest as the dataset size increases. This can be attributed to the limited number of data points per class, which can impact the performance of machine learning models.
We reckon that TabNet's poor performance over the HIPMP dataset could be due to the number of sample differences for each dataset. While DVM has a larger number of samples and HIPMP has samples under 2k which can make a complex transformer-based model like TabNet prone to overfitting.   

\subsection{Explainability}
\label{sec:shap}
\begin{figure}[h]
	\centering
	\subfloat[DVM]{\includegraphics[width=\linewidth]{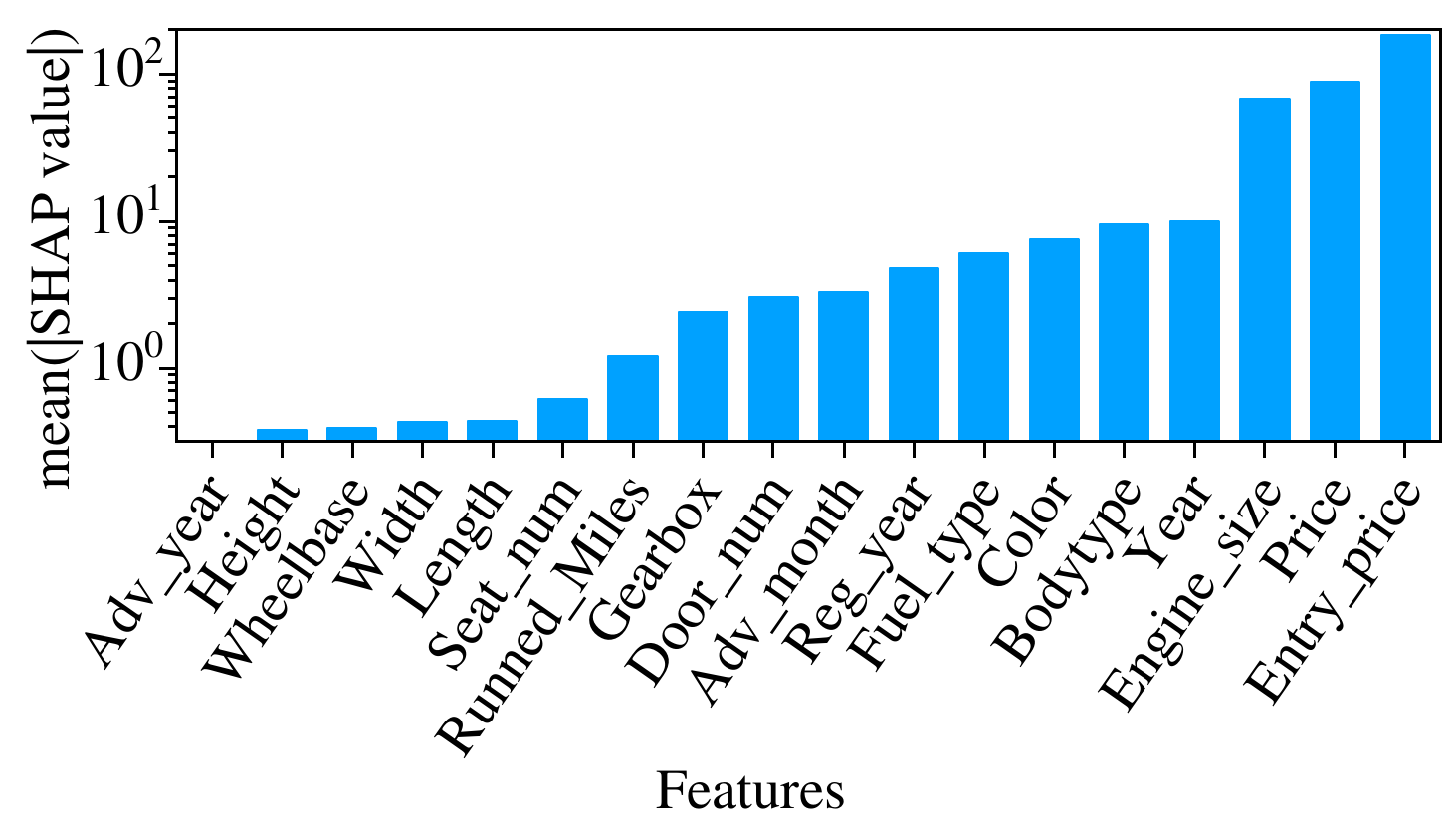}}\\
	\subfloat[HIPMP]{\includegraphics[width=\linewidth]{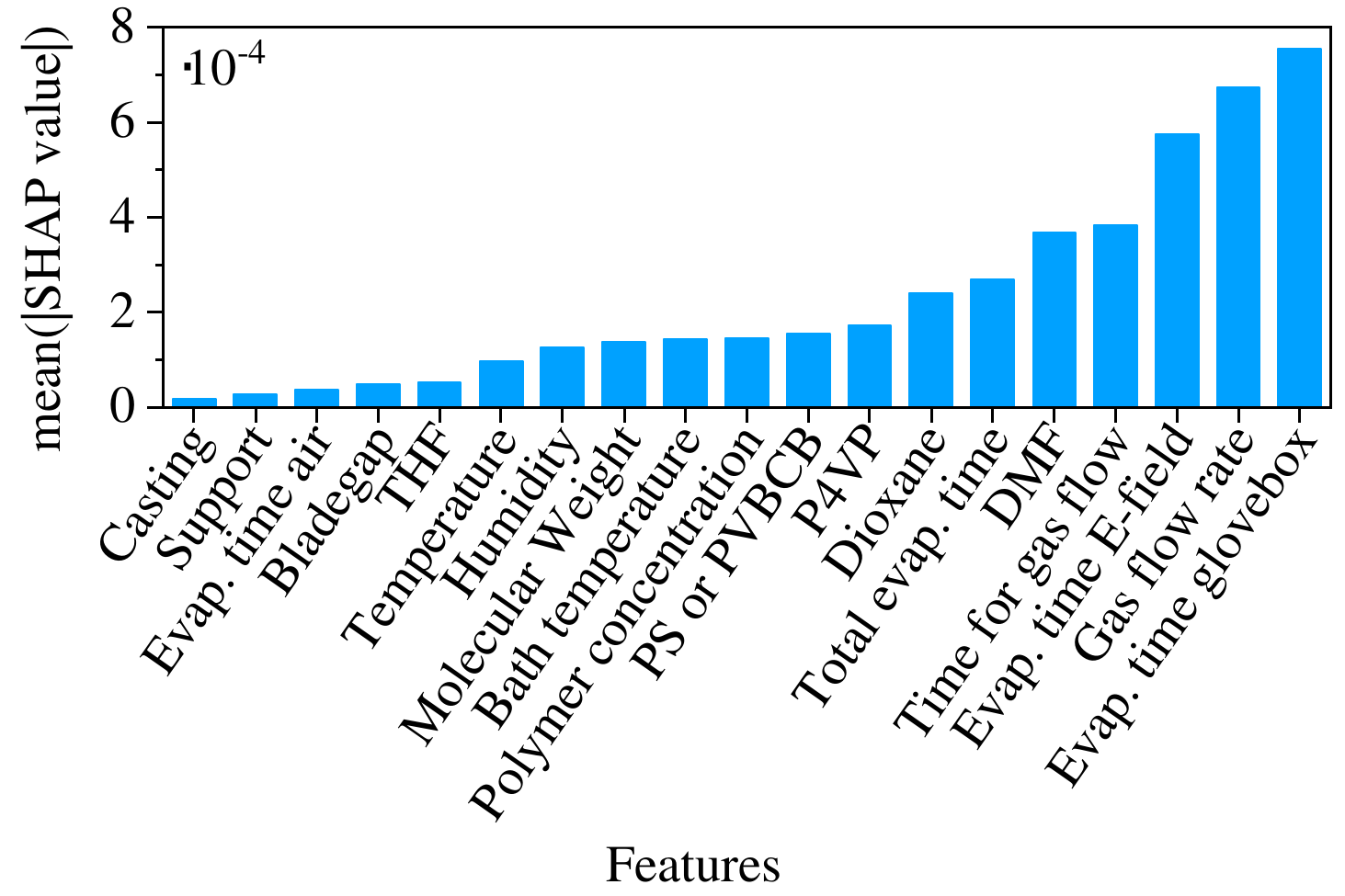}}
	\caption{Determining the influence of input features through the application of the SHAP method.}
	\label{fig:shap}
\end{figure}

Fig. \ref{fig:shap} visualizes the SHAP values associated with tabular embeddings for all input features, effectively highlighting the significance of tabular features, i.e. the input marketing parameters for the DVM dataset and the fabrication parameters for the HIPMP dataset. For the DVM dataset, the relatively high importance of the price, original price and engine size for predicting the car model is not surprising, as the values vary a lot and are usually quite specific for a certain model.

For the HIPMP dataset with polymer membranes, the prominence of the time for gas flow, evaporation time E-field, gas flow rate and evaporation time glovebox can be attributed to the relatively low frequency of non-zero values for these parameters and their well-known effect on the pore formation, as described in~\cite{Sankhala.2020, Dreyer.2019}. Of the frequently occurring fabrication parameters, the importance of the DMF content in the solvent mixture and the total evaporation time confirm the experimentalist's intuition. Similarly, the small effect of casting type and support material on the pore formation underlines the robustness of the fabrication process against changes that are disconnected from the membrane surface.
\section{Discussion and Conclusion}
\label{discussion}

In this paper, we introduced a novel approach: a multi-task multi-modal contrastive training strategy designed to enhance the performance of tabular models, especially in scenarios where direct access to image data isn't feasible at deployment. The latter is motivated by the fact that capturing scientific images is a labor- and resource-intensive process, whereas a wealth of information on fabrication parameters is available. Our solution involves pre-training on extensive datasets that combine tabular and imaging data,  aiming to leverage both tabular and imaging data for potentially improved inference performance with tabular data. We applied our approach to the demanding task of predicting the quality metrics of isoporous polymer membranes from microscopic images, surpassing all state-of-the-art tabular baselines, including our own proposed model. Furthermore, we demonstrated its versatility by successfully applying it to photographic images, excelling in predicting car models from advertisement data.

Our attribution and ablation study demonstrates that a multi-task training strategy, which combines a pretext task and contrastive learning, enhances the downstream task performance compared to a single-training strategy. We posit that jointly learning these two tasks improves the generalization capabilities of the tabular encoder involved, even with a small sample size. Furthermore, our method exhibits versatility, effectively handling various tasks, including regression and classification. The relative impact of each tabular input feature is given as a SHAP value diagram in the Sec. \ref{sec:shap}.

\textbf{Limitations}  The pre-training method discussed herein exhibits robust performance on the two datasets employed during the evaluation phase. Notably, these datasets originate from significantly distinct domains, which provides a preliminary indication of the method's versatility and adaptability. However, we acknowledge that the scope of this testing is not sufficiently extensive, due to a deficit in suitable public datasets, to unequivocally guarantee the external generalizability of this novel pre-training method.

\textbf{Conclusion} In summary, our study offers a straightforward and effective strategy for harnessing the combined power of tabular and imaging data within the framework of multi-task multi-modal contrastive learning. This approach holds particular relevance in scientific imaging, where we seek to capitalize on extensive multi-modal data during pretraining and apply it in unimodal contexts. We firmly advocate for the underexplored and often underestimated potential of tabular data in the realm of deep learning. It is both easily obtainable and ubiquitous, as it encompasses any numerical or categorical feature. Moreover, its intrinsic interpretability, with each feature directly representing a meaningful concept, is a valuable asset. We aspire to inspire future research to unlock the untapped possibilities that lie within this domain.

\section*{Acknowledgments}
The authors thank Helmholtz AI for funding this research within the COMPUTING project (grant number ZT-I-PF-5-135). The project team thanks Tomer Fried for his initial exploration of the HIPMP dataset. Thanks to all experimentalists producing membranes (Thomas Bucher, Meiling Wu, Zhenzhen Zhang, Christian Höhme, Kirti Sankhala, Muntazim Khan, Kristian Buhr, Oliver Dreyer, Sofia Rangou, Michael Appold, Benjamin Botev, Sarah Saleem, Erik Schneider, Lara Hub) and images (Clarissa Abetz, Anke Höhme, Erik Schneider, Evgeni Sperling). Thanks to Prof. Volker Abetz for scientific advice. Thanks to Anke Höhme for assistance in the dataset curation. Martin Held thanks Imagic Bildverarbeitung AG (Glattbrugg, Switzerland) for adapting the Image Imaging System (IMS) software for our data export requirements.

\bibliographystyle{IEEEtran}
\bibliography{main}

%
%
%
%
%
%
%
%

\clearpage


\section*{Supplementary material}

# Supporting Information - Code and Pixels: Multi-Modal Contrastive Pre-training for Enhanced Tabular Data Analysis

## I. 1D-RESNET-CBAM

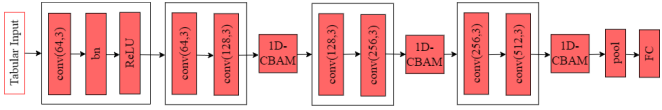

Fig. 1. The structure of our study's 1D-ResNet-CBAM.

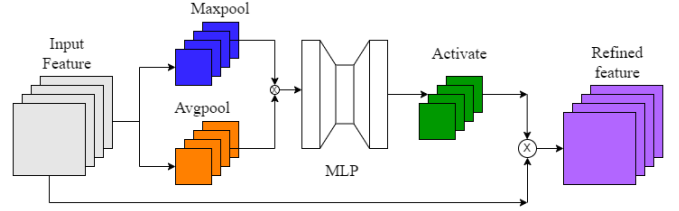

Fig. 2. The structure of our proposed 1D-CBAM, being integrated in the 1D-ResNet-CBAM model of Fig. 1.

### A. An Overview of 1D-ResNet-CBAM

As the convolutional neural network (CNN) delves deeper into its layers, there is an increased susceptibility to the degradation of gradients, ultimately leading to a decline in the model's overall performance. ResNet [1] has gained significant prominence in a variety of recognition tasks due to its utilization of identity mapping for shortcut connections, effectively mitigating the problem of gradient degradation [2]. Generally, deep convolutional neural networks possess a multitude of parameters and exhibit superior recognition accuracy when trained on extensive datasets. However, practical challenges often restrict the availability of sufficient data samples, resulting in limited dataset sizes. To address this data scarcity, a solution involves incorporating a 1D-convolutional block attention module (1D-CBAM) into the ResNet architecture. This augmentation enables the comprehensive extraction of pertinent features from individual tabular attributes, thereby enhancing the model's performance.

The structural configuration of the 1D-ResNet-CBAM network model introduced in this study is depicted in Fig. 1. The input to this model consists of tabular data, while the output comprises either a 33-dimensional membrane quality metric vector or, in the case of DVM, the specific class label associated with that particular tabular dataset. The convolutional layers within the 1D-ResNet-CBAM architecture employ $3 \times 3$ convolution kernels, adhering to three fundamental design principles: (1) To maintain consistency in the output feature map size, each layer employs an identical number of convolution kernels. (2) When the stride is set to 2, the convolution layer performs direct downsampling, reducing the size of the feature map by half while doubling the number of convolution kernels. This adjustment preserves the time complexity of each layer. (3) Following each convolution operation, batch normalization (BN) is applied to mitigate overfitting and prevent issues related to gradient vanishing or exploding within the network. Subsequently, the rectified linear unit (ReLU) activation function is utilized to maximize the utilization of gradient information, ensuring the model's continuous convergence.

In spite of the aforementioned benefits, there are specific challenges observed in the ResNet framework. Firstly, ResNet encounters limitations related to the size of convolution kernels. The dilemma arises from the fact that small kernels struggle to capture dependencies across disparate signal regions, whereas large kernels lead to increased parameter counts, potentially hindering training efficiency. Secondly, ResNet treats spatial and channel information with equal importance, which may not effectively capture crucial information. In response to these challenges, we introduce a novel approach, 1D-CBAM, aimed at enhancing the identification accuracy of ResNet. The proposed network structure incorporates 1D-CBAM following the final convolution layer, thereby enabling the fine-tuning of channel direction features. This adjustment serves to accentuate salient regions and extract more detailed features (Fig. 2).

In detail, the structure of the attention modules' 1D-CBAM, as depicted in Fig. 2, is designed to operate on an input feature map $F_{in} \in \mathcal{R}^{W \times C}$, where $W$ represents the length of the feature map, and $C$ signifies the number of channels. In the 1D-CBAM module, channel features are initially extracted by simultaneously applying max pooling and average pooling operations, and the resulting features are subsequently reshaped as follows:

$$c_1 = \text{MaxPool}(F_{in}); \max(F_{in}(1 \leq in \leq W, C)),$$
$$c_2 = \text{AveragePool}(F_{in}) = \frac{1}{W} \sum_i^W F_{in}(i, C), \quad (1)$$
$$c = \text{concat}(c_1, c_2)$$

Here, $c_1 \in \mathcal{R}^{1 \times C}$ and $c_2 \in \mathcal{R}^{1 \times C}$, and a concatenation operation is applied, followed by a nonlinear transformation carried out by a multilayer perceptron (MLP) consisting of two hidden layers, with the first hidden layer containing $C/r$ neurons, where $r$ denotes the reduction ratio. The output of the

TABLE I
OVERVIEW AND DESCRIPTION OF THE FABRICATION PARAMETER OF HIPMP DATASET USED AS INPUT TO THE TABULAR MODEL.

| Fabrication parameter | Definition |
| --- | --- |
| Molecular weight | Number-average molecular weight of the polymer in kg/mol |
| PS or PVBCB | Percentage of the molecular weight of the polystyrene or poly(4-vinylbenzocyclobutene) block, relative to the total molecular weight |
| P4VP | Percentage of the poly(4-vinylpyridine) block (sums to 100% with the above) |
| Polymer concentration | Concentration of the polymer in the solvent mixture in weight percent |
| DMF | Percentage of dimethylformamide in the solvent mixture |
| THF | Percentage of tetrahydrofurane in the solvent mixture |
| Dioxane | Percentage of dioxane in the solvent mixture (sums to 100% with the two above) |
| Casting | Type of casting method: manual or machine (treated as categorical) |
| Support | Type of support material, on which the polymer is cast upon: Different types of nonwoven or glass (treated as categorical) |
| Blade gap | Gap between the casting blade and the surface of the support material in micrometers |
| Evaporation time air | Time of the cast solution after passing the blade and before immersing into the water precipitation bath, allowing the solvent mixture to partially evaporate into air, in seconds |
| Evaporation time glovebox | Identical to "evaporation time air", but in a dry nitrogen atmosphere instead of air |
| Evaporation time E-field | Identical to "evaporation time air", but in a setup for applying an electric field, with the electric field being turned off (0 kilovolt) |
| Time for gas flow | Identical to "evaporation time air", but in a setup for applying a gas flow |
| Gas flow rate | Rate of gas flow in milliliter per minute |
| Total evaporation time | Sum of all aforementioned evaporation and gas flow times |
| Bath temperature | Temperature of the water precipitation bath, in degree Celsius |
| Humidity | Humidity of the air or nitrogen atmosphere, in percent |
| Temperature | Temperature of the air or nitrogen atmosphere, in degree Celsius |

channel attention mechanism is obtained through the sigmoid activation function:

$$Out_c = \text{sigmoid}(W_{MLP}(c)) \quad (2)$$

In this equation, $W_{MLP}$ represents the weights of the MLP. The channel attention mechanism learns a vector, $Out_c$, with a length of $C$, indicating the weight values assigned to each channel to assess the importance of different channel features. The final output feature map, $F_{out}$, is obtained by element-wise multiplication (Hadamard product) between $Out_c$ and $F_{in}$. Channels containing more valuable information receive higher weightings, as expressed by:

$$F_{out} = F_{in} \otimes Out_c \quad (3)$$

## II. HEREON ISOPOROUS POLYMER MEMBRANE PRODUCTION (HIPMP) DATASET

### A. Overview of the Hereon Isoporous Polymer Membrane Production (HIPMP) Dataset

When fabricating isoporous flat sheet membranes from block copolymers, the settings of the production process still comprise trial-and-error-based choices, leading to a waste of time and material. To address this challenge, the Hereon Isoporous Polymer Membrane Production (HIPMP) dataset enables relating the fabrication parameters with the morphological properties of the resulting membrane, which serve as membrane quality metrics, in order to investigate the membrane fabrication process.

One tabular component of the HIPMP dataset encompasses the fabrication parameters of PS-P4VP block copolymer membranes, as recorded by the experimentalist when using the in-house-built membrane casting machine or when manually performing the membrane casting process [3]–[17]. While a total of 19 fabrication parameters are documented, not all of them can be controlled. Rangou *et al.* provide a detailed description of the fabrication process [11], while Dreyer *et al.* [5] and Sankhala *et al.* [13] covered the addition of electric field and gas flow, respectively. We included only samples with an electric field of 0 kV or no field at all. However, we included gas flow rates up to 6000 mL/min, as this was found not to affect the surface pore morphology [13]. The polymer PVBCB-P4VP is chemically slightly different from PS-P4VP, but its membranes are fabricated in the same manner [12]. Other membranes with chemical post-modification (e.g. chemical functionalization with molecules, deposition of oxides via atomic layer deposition), physical post-treatment (e.g. filling pores with liquid metal, membranes after water flux or porometer analysis, membranes after fouling tests), fabrication methods other than blade casting (e.g. spray-coating, roller coating, dip coating) or alterations of the blade casting method (e.g. evaporation in solvent atmosphere, employing additives to the casting solution, evaporation in electric fields > 0 kV) were explicitly excluded. Post-treatment with UV-light, and optional follow-up temperature treatment, was not excluded, as this procedure did not alter the surface pore morphology.

In brief, the fabrication of these membranes involves spreading a block copolymer solution onto a nonwoven using a coating blade, resulting in a membrane thickness determined by the solution's concentration and the blade gap. The combination

of evaporation-induced self-assembly and nonsolvent-induced phase separation creates an asymmetric layered architecture, with a spongy support layer and an isoporous top layer integrated into a single flat sheet membrane. A list with a short description of each parameter is available in Tab. I. Here the entries of the categorical parameter Casting were reduced to "machine" or "manual". Similarly, the entries for Support were aggregated to nonwoven, PAN or glass. Empty values in this tabular dataset were either filled with 0 (Evaporation time glovebox, Evaporation time E-field, Time for gas flow, Gas flow rate) or imputed with the mean of the existing values (Molecular weight, PS or PVBCB, P4VP, Polymer concentration, DMF, THF, Casting, Support, Blade gap, Evaporation time air, Bath temperature, Humidity, Temperature), while the values for the Total evaporation time and Dioxane were calculated, as indicated in Tab. I. Subsequently, the mean and standard deviation of each parameter were normalized to 0 and 1, respectively.

The second basis of the dataset is a comprehensive collection of 1970 microscopic images of the porous surface of PS-P4VP copolymer membranes. The images were recorded by scanning electron microscopy, either with a Zeiss Merlin or a Zeiss Leo model, employing an Inlens detector (without signal mixing with other detectors) at 50kx magnification (1 pixel = 2.233 nm). Other magnifications and detectors were excluded, as well as tilted images, images of the bottom surface of membranes and images with measurement lines. These grayscale images all share a uniform size of $1024 \times 768$ pixels, from which a bar of $1024 \times 60$ pixels with a microscopy label is cropped at the images' bottom, resulting in a $1024 \times 708$ sized images. Visualized examples are provided in Fig. 4a and Fig. 5a. To derive morphological membrane properties, a two-stage approach was devised. Initially, characteristic morphology information was extracted from the images using a deep-learning segmentation model. Subsequently, in the second stage, 47 morphological quality properties were generated from the model's output. Both stages are elaborated upon in the subsequent sections.

### B. Segmentation

Given the essential importance of spatial information, the task of extracting morphological information is framed and interpreted as a segmentation problem. The initial phase involved the definition of classes relevant to the membrane quality and performance. This is done by domain experts. Tab. II provides an overview and concise descriptions of these classes. Additionally, a background class was established to encompass areas outside the specified classes of interest. Despite the interest in individual pore instances, semantic segmentation is utilized instead of instance segmentation, primarily due to the distinct and non-overlapping nature of the pores within the dataset. The pores exhibit clear separation from one another, eliminating the need for instance-level differentiation in this stage. This strategic choice not only leveraged the benefits of semantic segmentation tailored to our specific application but also reduced unnecessary complexity during the labeling and training process.

TABLE II
OVERVIEW AND DESCRIPTION OF THE CLASSES (LABELS) USED TO DEFINE THE SEGMENTATION TASK.

| Class Name | Definition |
| --- | --- |
| Covered Area | Large area ($\sim$>100nm²) with no visible pore. |
| Open Pores | Hollow cylindrical structure that traverses the isoporous layer of the membrane entirely in a vertical orientation. |
| Closed Pores | Closed cylindrical structure that is clearly visible but does create an opening in the membrane surface. |
| Internal Structures | Non-cylindrical opening breaking through the membrane surface. |
| Artifact Blob | Bright circular spots on the membrane surface of unknown origin. |
| Artifact Dust | Dust/dirt/contamination material lying on the membrane surface. |

The chosen semantic segmentation model was nnUNet [18], due to its state-of-the-art performance across different domains, which instilled confidence in its ability to deliver accurate and robust segmentation results for electron microscopy images. To efficiently generate semantic segmentation masks for the entire dataset, a human-in-the-loop active learning strategy was employed. In an iterative process, human experts identify the most informative data samples, which are first labeled and then added to the training data on which the deep learning model was trained. By analyzing the model predictions, failure cases were identified, prompting the selection of new samples for inclusion in the training pool. This iterative process continued until the model's prediction quality met the criteria set by human experts. During the initial iteration, labeling was done using a random forest approach complemented by human corrections. Subsequent iterations replaced the random forest with predictions from the model of the previous iteration. Again, the refinement and correction of segmentation masks were performed by human experts. A total of three iterations ensued, resulting in 28, 42, and 65 labeled images in their respective iterations.

Afterward, the final model was used to predict the entire dataset and generate segmentation masks for all images, an example is shown in Fig. 4. Statistics about the class distribution in the resulting dataset as well as for each active learning iteration are presented in Fig. 3.

### C. Parameter Extraction

In the subsequent phase, the segmentation masks acquired earlier serve as the basis for generating morphology membrane properties. These properties are devised to characterize both the functionality of the membrane and to provide insights into the membrane manufacturing process. They can be divided into three distinct categories: general class attributes, pore-specific details, and pattern characteristics. The classes of interest align with those previously defined for the segmentation task, with the addition of an 'All Pores' class, representing the union of open and closed pores. Consequently, these 7

3

TABLE III
LIST OF ALL MEMBRANE QUALITY PARAMETERS, DERIVED FROM THE SEGMENTATION MASKS. CLASS INSTANCES ARE IDENTIFIED USING A CONNECTED COMPONENTS ALGORITHM. THE GENERAL CLASS INFORMATION IS GENERATED FOR EACH OF THE 7 CLASSES WHICH RESULTS IN 28 QUALITY PARAMETERS. THE PORE INFORMATION IS ONLY COMPUTED FOR THE 3 PORE CLASSES, RESULTING IN 12 PARAMETERS. PATTERN INFORMATION IS REPRESENTED BY A TRIANGLE MESH GENERATED THROUGH DELAUNAY TRIANGULATION AMONG ALL PORE INSTANCES. TOGETHER WITH THE 7 PARAMETERS DESCRIBING THE PATTERN INFORMATION, A TOTAL OF 47 PARAMETERS IS OBTAINED.

|  | Property | Description |
|---|---|---|
| General Class Information | Covered Area | Count of all pixels in the image that belongs to the class (converted into nm²). |
|  | Number of Instances | Count of not connected instances of each class. |
|  | Size of Instances | Mean and standard deviation of the number of pixels belonging to individual class instances (in nm²). |
| Pore Information | Diameter | Mean and standard deviation of the diameters (in nm) of each instance belonging to the class. The diameter is defined as the diameter of a circle with the same area as the pore. |
|  | Circularity | Mean and standard deviation of the circularity of each instance belonging to the class. The circularity of an instance is computed by the ratio of the length of the minor axis and the length of the major axis of the pore instance. |
| Pattern Information | Pore Distance | Mean and standard deviation of all edge lengths (in nm) in the triangle mesh. |
|  | Pore Connectivity | Number of edges in the triangle mesh. |
|  | Pattern Distribution | Mean and standard deviation of the size of all triangles (in nm²) in the triangle mesh. |
|  | Pattern Regularity | Mean and standard deviation of the regularities of all triangles in the triangle mesh. The regularity of a single triangle is computed by the ratio of the shortest and the longest edge of the triangle. |

classes collectively are the basis for defining the membrane properties. For an overview and concise descriptions of all these properties, refer to Tab. III. The general class information encompasses 4 individual properties, each quantifying the presence or absence of one of the 7 classes. Pore-specific information is exclusively derived from the 3 pore-related classes and encompasses 4 distinct parameters. These parameters are specific to individual pores and are aggregated across the entire image, offering insights into the functionality of individual pores and the complete membrane. Pattern information is extracted from the 'All Pores' class to depict the distribution pattern of pores across the membrane. This results in an additional 7 properties, contributing to a grand total of 47 membrane quality properties. A selection of these is visualized in Fig. 5.

*D. Public Version of the Hereon Isoporous Polymer Membrane Production (HIPMP) Dataset*

A portion of 479 images of the HIPMP dataset used for the MT-CMTM model in this work corresponds to previously unpublished fabrication recipes. Due to protection of intellectual property, we created a public version of the HIPMP dataset [19], which was reduced to 1491 images. In the tabular data, each image is assigned to the DOIs of the publications containing the corresponding recipe.

However, the analyzed polymer properties (PS or PVBCB; molecular weight) are often refined by the experimentalist in parallel to the membrane production, leading to numerical changes of up to $\pm 5\%$ for the PS or PVBCB percentage (and thus P4VP percentage) and up to $\pm 15$ kg/mol for the molecular weight. Similarly, also fabrication parameters summarized in the publications might differ from individual samples as a result of a compact statement in the paper. Thus, the total evaporation time may be off by $\pm 3$ s (evaporation time air, glovebox, E-field and time for gas flow are contained therein), and the polymer concentration by $\pm 5$ w%. The percentage of DMF, THF, dioxane and the bladegap are considered to be reported precisely. The casting type, support material, gas flow rate, bath temperature, temperature and humidity are disregarded for matching the tabular fabrication parameters with literature fabrication recipes, as they are not always reported.

The following procedure was applied to match the fabrication parameter set of each HIPMP dataset row, i.e. each image, with the fabrication recipes from literature. If there is a full match of molecular weight, PS or PVBCB, DMF, THF, total evaporation time, polymer concentration and bladegap, the DOI is directly assigned to the image. Without a full match, the HIPMP molecular weight is expanded by $\pm 15$ kg/mol and compared to the literature values, filtering all matches. Of these matches, the HIPMP's "PS or PVBCB" value is checked for a direct match, otherwise expanded by $\pm 5\%$ to filter matches. Of these, the "DMF" values are filtered for direct matches. Of these, the "THF" values are filtered for direct matches. Of these, the total evaporation time is checked for a direct match, otherwise expanded by $\pm 3$ s to filter matches. Of these, the polymer concentration is checked for a direct match, otherwise expanded by $\pm 5$ w% to filter matches. Of these, the bladegap is filtered for direct matches. Whenever there is no match, the loop is exited and the fabrication parameter set is considered unpublished.

In summary, the HIPMP dataset comprises both image and tabular data, enabling a comprehensive analysis of the membrane fabrication process. The dataset contains 1970 or 1491 grayscale images of PS-P4VP copolymer membranes, for this publication's model or in the public dataset version, respectively. These encompassed images underwent a random crop to

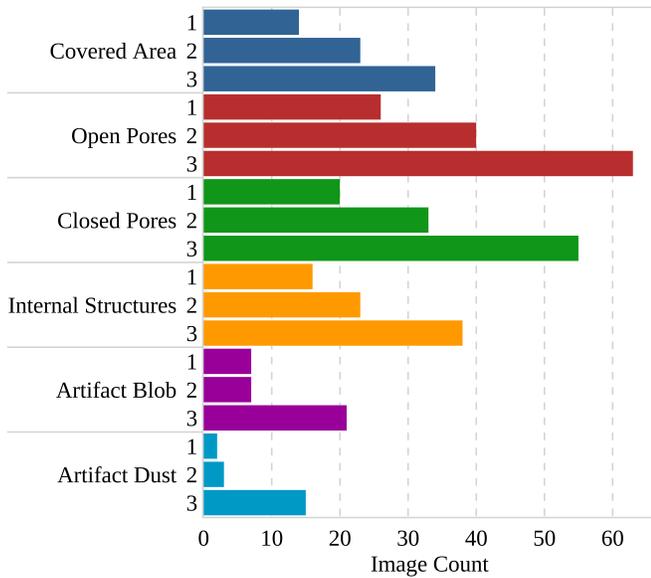

(a) Each iteration in the active learning pipeline

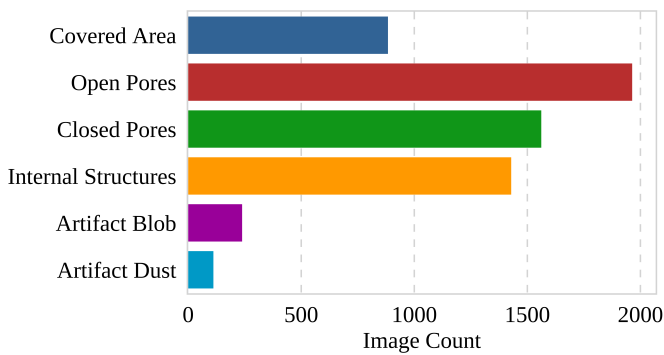

(b) Complete HIPMP dataset

Fig. 3. Class occurrences. (a) Number of images containing each class for each iteration (1, 2, 3) in the active learning pipeline. In total 28, 42, and 65 images were annotated in the corresponding iteration. (b) Number of images containing each class in the final dataset which comprises 1970 images.

$224 \times 224$ pixels and a random horizontal flip as augmentation methods, before inputting them into the MT-CMTM pipeline. Each image is characterized by 19 fabrication parameters and 47 membrane morphological quality properties. Out of the latter, only 33 non-sparse quality properties were used for training the machine learning models. Sparse properties were merely present on very few images, e.g. artifact dust size mean and standard deviation or artifact blob size mean and standard deviation. In order to find the best-performing model for the prediction of all non-sparse properties *en masse*, the corresponding 33 MSE values were averaged via the sklearn function.

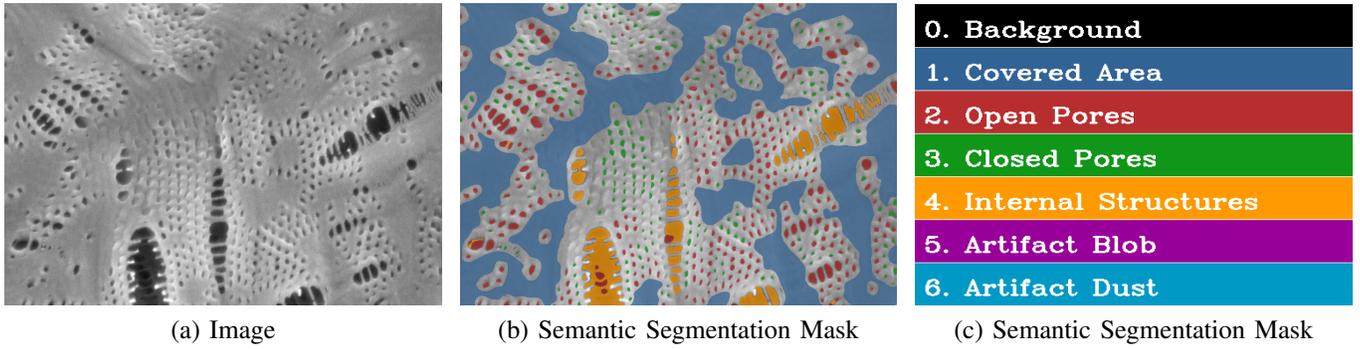

Fig. 4. Image of (a) a PS-P4VP copolymer membrane together with (b) the segmentation mask generated by the deep learning semantic segmentation model.

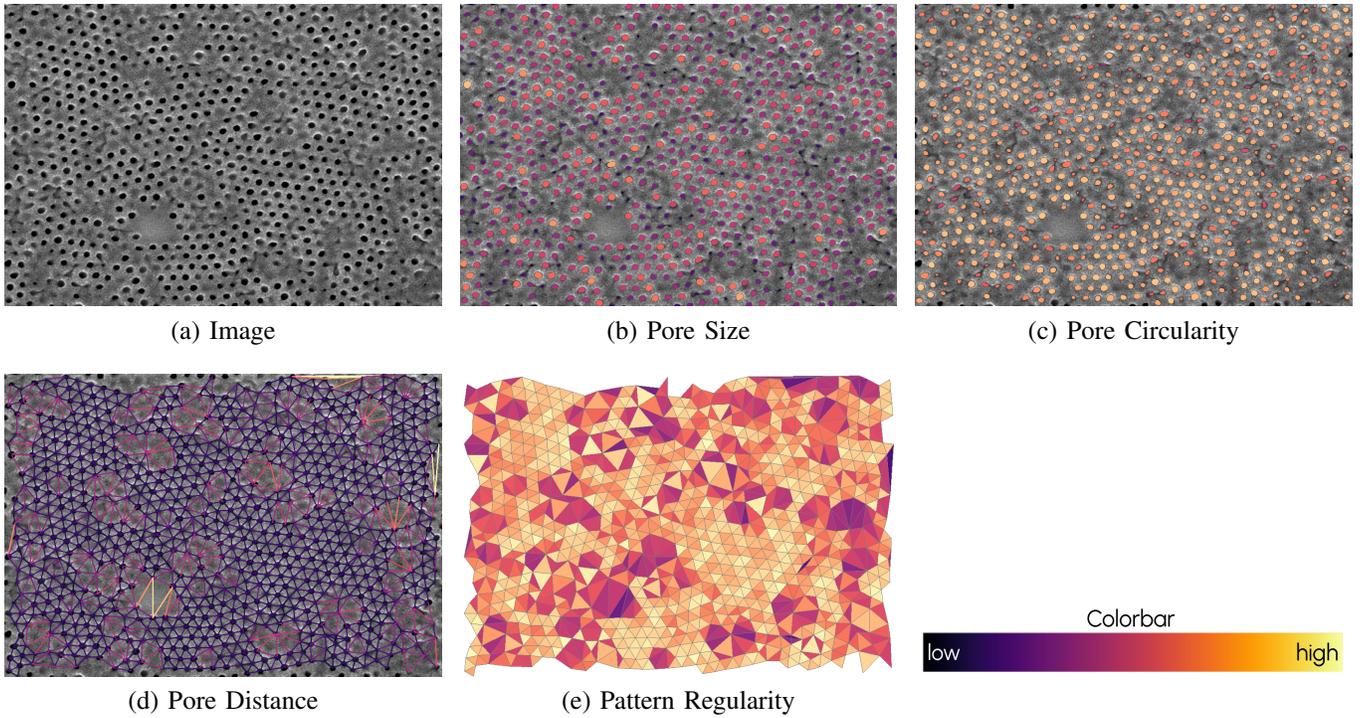

Fig. 5. Image of a PS-P4VP copolymer membrane together with the visualizations for a selection of topological membrane properties.



\end{document}